\documentclass{article}

\usepackage{arxiv}

\usepackage[utf8]{inputenc} 
\usepackage[T1]{fontenc}    
\usepackage{hyperref}       
\usepackage{url}            
\usepackage{booktabs}       
\usepackage{amsfonts}       
\usepackage{nicefrac}       
\usepackage{microtype}      
\usepackage{lipsum}
\usepackage{graphicx}
\graphicspath{ {./images/} }

\title{Deep Learning-Driven Microstructure Characterization and Vickers Hardness Prediction of Mg-Gd Alloys}

\author{
 Lu Wang \\
  School of Computer Engineering and Science\\
  Shanghai University\\
  Shanghai, China 200444 \\
  \and
 Hongchan Chen \\
  School of Computer Engineering and Science\\
  Shanghai University\\
  Shanghai, China 200444 \\
  \and
 Bing Wang \\
  School of Computing and Information\\
  University of Pittsburgh\\
  Pittsburgh, PA 15213 \\
  \and
 Qian Li \\
  School of Computer Engineering and Science\\
  Shanghai University\\
  Shanghai, China 200444 \\
  \and
 Qun Luo\thanks{\texttt{qunluo@shu.edu.cn}} \\
  School of Computer Engineering and Science\\
  Shanghai University\\
  Shanghai, China 200444 \\
  \and
 Yuexing Han\thanks{\texttt{han\_yx@i.shu.edu.cn}}  \\
  School of Computer Engineering and Science\\
  Shanghai University\\
  Shanghai, China 200444 \\
}

\begin{document}
\maketitle
\begin{abstract}
In the field of materials science, exploring the relationship between composition, microstructure, and properties has long been a critical research focus. The mechanical performance of solid-solution Mg-Gd alloys is significantly influenced by Gd content, dendritic structures, and the presence of secondary phases. To better analyze and predict the impact of these factors, this study proposes a multimodal fusion learning framework based on image processing and deep learning techniques. This framework integrates both elemental composition and microstructural features to accurately predict the Vickers hardness of solid-solution Mg-Gd alloys. Initially, deep learning methods were employed to extract microstructural information from a variety of solid-solution Mg-Gd alloy images obtained from literature and experiments. This provided precise grain size and secondary phase microstructural features for performance prediction tasks. Subsequently, these quantitative analysis results were combined with Gd content information to construct a performance prediction dataset. Finally, a regression model based on the Transformer architecture was used to predict the Vickers hardness of Mg-Gd alloys. The experimental results indicate that the Transformer model performs best in terms of prediction accuracy, achieving an $R^2$ value of 0.9. Additionally, SHAP analysis identified critical values for four key features affecting the Vickers hardness of Mg-Gd alloys, providing valuable guidance for alloy design. These findings not only enhance the understanding of alloy performance but also offer theoretical support for future material design and optimization.
\end{abstract}


\section{Introduction}
\label{sec:introduction}
Magnesium alloys, with high specific strength, excellent casting properties, vibration-damping and noise-reduction capabilities, and ease of processing, show great potential for application in various commercial sectors. In the automotive, electronics, and aerospace industries, the alloys have gradually became a strong alternative to aluminum alloys, steel, and engineering plastics \cite{JOOST2017107, KIANI201499, YANG2021705}. In the automotive industry, the lightweight advantage of magnesium alloys has made them widely used in the manufacturing of engine components, body structures, and wheels, helping to improve fuel efficiency and reduce the overall weight of vehicles. In the aerospace sector, magnesium alloys, with high specific strength and good fatigue resistance, have become ideal materials for aircraft structural components and other critical parts, driving the design and production of lightweight airplanes.

In recent years, researchers have further optimized the properties of magnesium alloys by adding rare earth elements (such as Gd, Y, and Er), significantly improving their strength and wear resistance \cite{nie2012precipitation}. The addition of rare earth elements can form strengthening phases, thereby enhancing the mechanical properties of the material \cite{TANG2019172, HE2007316, chu2021microstructure}. For instance, He et al. \cite{HE2007316} discovered that by optimizing heat treatment parameters, cast Mg-10Gd-2Y-0.5Zr alloys achieved a good combination of high strength and adequate ductility at room temperature, with a tensile strength of 362 MPa, a yield strength of 239 MPa, and an elongation of 4.7$\%$. Wang et al. \cite{WANG2022111658} reported that Mg-Gd-Y-Zr alloys, produced by extrusion casting and heat treatment, reached a tensile strength of 383 MPa at room temperature or and at 200°C. Among all rare earth elements, Gd is widely considered a key element for improving the performance of magnesium alloys, as its introduction effectively enhances the strength and wear resistance of the material. However, as the composition and microstructure complexity of magnesium-rare earth alloys increase, traditional performance testing methods face challenges. Vickers hardness (HV) is an important indicator for measuring material hardness, reflecting the ability of magnesium alloys to resist deformation and wear. The higher the hardness, the less likely the material will undergo plastic deformation under external forces, which is crucial for structural components in the automotive and aerospace industries that are often subjected to high stress and high wear environments. To efficiently predict the Vickers hardness of Mg-Gd alloys, researchers have begun to explore new methods. Traditional Vickers hardness measurements typically rely on experiments or use density functional theory (DFT) to calculate bulk modulus and shear modulus, which are often time-consuming and costly \cite{SALEHI2022121663, ZHANG2022121566}. To address these issues, our research introduces deep learning and image processing techniques to automate the analysis of the microstructure of magnesium-rare earth alloys. By establishing a correlation model between microstructural features and Vickers hardness, we can accurately predict the hardness properties of the alloys. This method reduces human intervention in the experimental process, improves research efficiency, and accelerates the optimization of alloy performance, achieving significant progress especially in predicting and enhancing Vickers hardness.

Segmentation methods for metallographic images can be classified into rule-based approaches and learning-based approaches. Rule-based techniques integrate traditional image processing methods with domain-specific knowledge, allowing the extraction and analysis of crucial information from material images based on the inherent characteristics of each image type. Research indicates that extracting features of different phases from microscopic images facilitates more convenient analysis. However, these methods necessitate the manual definition of multiple parameters, a process that is not only time-consuming but also presents challenges in determining accurate parameter values. For example, Liveres and Pilkey applied the minimum cross-entropy method and cost function to determine an adaptive threshold for the second-stage segmentation \cite{LIEVERS2004134}. Dutta et al. employed Otsu for segmenting the microstructure of dual-phase steel, followed by extracting statistical features to identify morphologies after different heat treatments \cite{DUTTA2019595}. On the other hand, some scholars combined supervised machine learning to propose learning-based methods. Classifiers distinguish microstructural characteristics in images by learning extracted features. Generally, learning-based methods outperform rule-based ones. For metallographic images, the feature space mainly includes texture, grayscale values, spatial relationships, corners, and morphology. Image segmentation is achieved by extracting morphological features of microstructures using learning-based methods \cite{Han2022CenterenvironmentFM}. Specifically, Gola et al. evaluated second-phase particles in dual-phase steel using threshold segmentation based on morphological features and Support Vector Machine (SVM) \cite{GOLA2018324}. Choudhury et al. combined Watershed with a Convolutional Neural Network (CNN) to achieve phase recognition in steel \cite{Choudhury2019ComputerVA}. Chen et al. proposed hierarchical parameter transfer learning for the segmentation of Mg2Si and Fe microstructures in aluminum alloys \cite{Chen2019AluminumAM}. Ma et al. based on DeepLab introduced symmetric overlapping tiling and three-dimensional symmetric rectification for segmenting dendrites and eutectic phases \cite{sym10040107}. Han et al. employed the superpixel technique for feature extraction from images and integrated it with DenseNet to achieve segmentation of high-resolution complex texture material images in a few-shot scenario \cite{Han2022RecognitionAS}. 

The crystal boundary detection problem in metallographic images is similar to but not entirely the same as the image segmentation problem. Image segmentation primarily involves separating foreground objects from the underlying background, emphasizing differences in various microstructures. For crystal boundary detection tasks, besides focusing on differences in various microstructures in the image, more attention needs to be paid to the boundaries of microstructures, namely the lines between grains. This emphasis on boundary characteristics makes crystal boundary detection somewhat more complex, requiring special methods to capture and analyze these small yet crucial features. This is essential for understanding the crystal structure and properties of materials. Related research, such as zhenying et al. developed an improved mean-shift strategy for boundary detection, addressing the challenge of blurred or disconnected boundaries \cite{Zhenying_2018}. Gorsevski et al. utilized two-dimensional cell automatic boundary extraction \cite{GORSEVSKI2012136}. Peregrina et al. proposed a preprocessing module comprising image simplification, noise reduction, automatic thresholding, and grain subdivision for subsequent measurement of grain size in low-carbon steel \cite{PEREGRINABARRETO2013249}. Ma et al. developed an improved graph cut method incorporating three-dimensional information for detecting missing boundaries \cite{MA20195}. Paredes et al. employed a watershed + labeling method based on the final opening advanced correlation function for measuring grain size \cite{PAREDESORTA2019193}. Lu et al. utilized the Level Set Method (LSM) to obtain closed boundaries through iterative solution of the level set equation \cite{LU2009267}. Additionally, learning-based methods such as Neural Networks (NN) \cite{DENGIZ2005854} and Support Vector Regression (SVR) \cite{Gajalakshmi2017GrainSM} have also been used for grain boundary measurement. Mingchun Li et al. proposed a more comprehensive Convolutional Feature (RCF) architecture based on multi-task learning for grain boundary detection and segmentation of metallographic images \cite{Li2020GrainBD}. 

\begin{figure*}[!htp]
  \centering
  \includegraphics[width=6in]{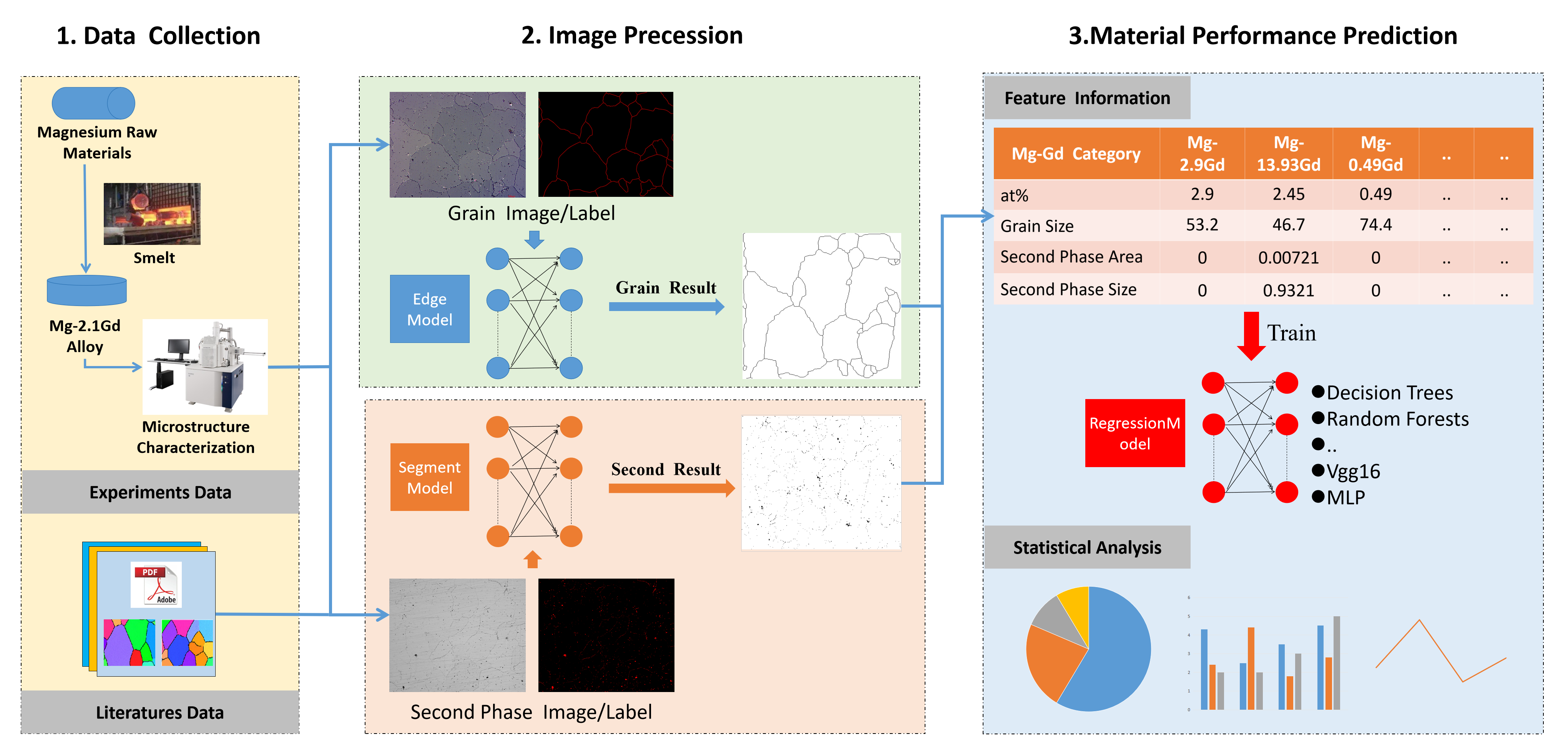}
  \caption{Overall framework of this study (revealing the relationship between material images, material microstructure, and material properties).}
  \label{fig:allModel1}
\end{figure*}

We develop a comprehensive research model that spans from image analysis to microstructural characterization and material performance. This study is primarily driven by deep learning, supported by computer vision techniques and a quantitative approach to material performance. Addressing the specific characteristics in metallographic images, this paper introduces a multitask automated image segmentation model within the realm of computer science. The model includes a deep learning-based multi-stage metallographic grain boundary detection model and a $GdH_2$ second-phase extraction model. The grain boundary extraction model primarily employs pixel-wise difference convolution to extract alloy grain boundary features. This convolutional kernel seamlessly integrates traditional edge detection operators into a conventional convolutional neural network, facilitating the direct extraction of edge features from images. Moreover, in the second stage of the model, the gradient information of the image is again utilized to construct an image reconstruction mask, mitigating edge loss issues during the grain boundary detection phase and ultimately enhancing the overall performance of the grain boundary detection task. For the second-phase metallographic images, a deep learning-based multiscale fusion segmentation model is utilized to extract the microstructure of the $GdH_2$ second phase. Subsequently, the average size of Mg alloy grains and the average area and equivalent circular diameter of the second phase are computed based on the extracted grain boundaries and $GdH_2$ microstructure. Finally, using 21 sets of microstructural information of Mg-Gd alloys combined with the corresponding Gd content data, we successfully predict the Vickers hardness (HV) of Mg-Gd alloys with a Transformer-based deep learning model \cite{vaswani2017attention}, achieving a coefficient of determination ($R^2$) of 0.9. The specific overall framework is shown in Figure \ref{fig:allModel1}.

The remainder of the paper is organized as follows: In Section \ref{sec:materials}, we have introduced the data used in this method, the image segmentation techniques employed, and the prediction methods for the material properties of Mg-Gd alloys. In Section \ref{sec:experiments}, we verify the effectiveness of our proposed method through Mg-Gd alloy images. Then, the results are also analyzed and discussed. Finally, the conclusion and the future works are described in Section \ref{sec:conclusion}.

\section{Dataset and Methods}
\label{sec:materials}

The section will provide a detailed introduction to the Mg-Gd alloy dataset used in this work and the microstructural information model for predicting the Vickers hardness of Mg-Gd alloys. The main content includes the Mg-Gd dendrite detection method, the $GdH_2$ second phase extraction method, and the prediction method for the Vickers hardness of Mg-Gd alloys.

\subsection{Mg-Gd Alloy Dataset}
\label{sec:alloy}

In the work, we aim to explore the relationship among composition, microstructure, and material properties by constructing a multimodal fusion learning framework. This framework predicts the Vickers hardness of Mg-Gd alloys by combining the atomic percentage of Gd with the microstructural characteristics of the alloy. Mg-Gd alloys mainly consist of two different scales of microstructures: dendritic structures in the magnesium matrix and elongated or point-like second-phase structures. To study the relationship between Gd content in the matrix, microstructural features, and the performance of Mg-Gd alloys, we built a performance prediction dataset that includes both experimental data and relevant literature data.

For the experimental data, Mg-2.1Gd alloys were melted using pure magnesium and Mg-35.9Gd master alloy. The specific process is as follows: Gd loss of 14 wt.$\%$ was added to the raw materials, and the surface oxidation layer was polished with 50-mesh sandpaper. The prepared mixture was preheated and then placed in a resistance furnace. Pure magnesium was fully melted under SF6+CO2 protective gas. Then, Mg-Gd alloy was added, maintaining the temperature and skimming off the slag. Finally, the molten alloy was poured into preheated cast iron molds. The resulting ingot was identified as Mg-2.1Gd (at.$\%$) through ICP-AES analysis. Subsequently, microstructural images of the samples were acquired. However, during the sample preparation process, a 5$\%$ nitric acid and alcohol solution was required for etching. Adjacent grains may exhibit similar crystallographic orientations, leading to uniform corrosion along the grain boundaries and making it difficult to distinguish them. Although extending the corrosion time can improve the clarity of the grain boundaries, it may also result in the precipitation of more $GdH_2$ second phases at the grain boundaries and within the grains, increasing the complexity of observation and further complicating the microstructural analysis.

Furthermore, to compensate for the lack of experimental data and improve the model performance, we collected microstructural images and performance data of Mg-Gd alloys with different Gd contents from relevant literature to supplement our dataset. Next, deep learning models were used to extract grain boundaries and second-phase microstructures of various Mg-Gd alloys. Subsequently, the Gd atomic percentage, grain size, second-phase area fraction, and particle size were used as feature data to predict the Vickers hardness of the alloy using machine learning and deep learning models. Additionally, to gain deeper insights into the "black box" of the machine learning models, feature importance analysis and SHAP (Shapley Additive Explanations) analysis were applied to interpret how each feature affects the target mechanical performance.

\subsection{Identification of Microstructures of Mg-Gd}
\label{sec:approach}

In the section, a detailed analysis method for the microstructure of magnesium alloys is introduced, mainly including Mg-Gd dendrite detection method and $GdH_2$ second phase extraction method.

\subsubsection{Mg-Gd Dendrite Detection Method}

\noindent\textbf{Edge Detection}

\begin{figure*}[!htp]
  \centering
  \includegraphics[width=6in]{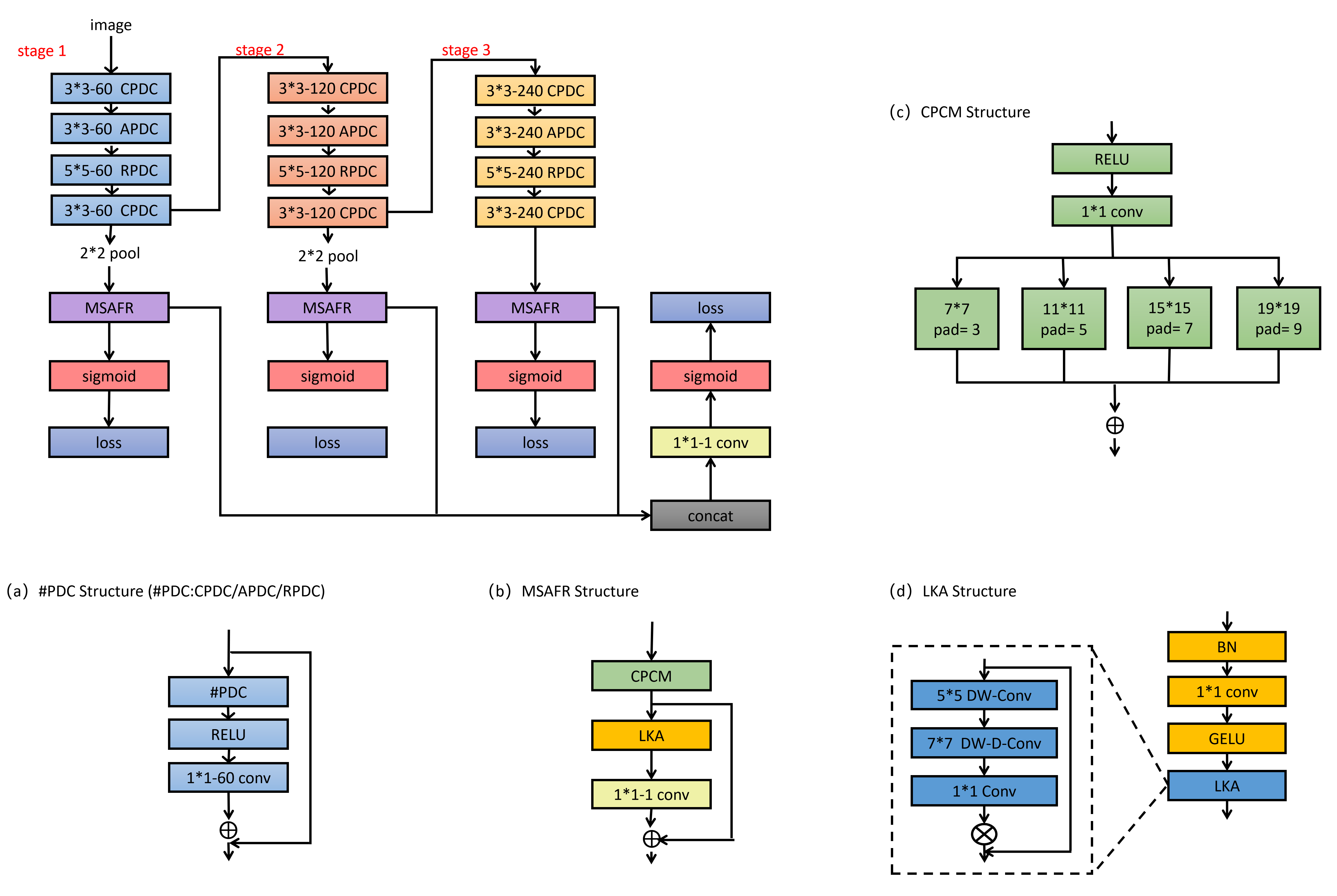}
  \caption{Edge detection model structure. The entire model framework consists of three stages. (a) presents the general structure of the edge extraction module; (b) is multi-scale adaptive feature refinement module; (c) is based on the compact part convolution module; (d) is based on channel and spatial large-kernel attention module.}
  \label{fig:process}
\end{figure*}

This study focuses on the image processing of solution-treated Mg-Gd alloys, where phases are segmented by detecting grain boundaries within each crystal. We developed a deep learning-based grain boundary detection model that primarily utilizes pixel difference convolution to identify the majority of Mg-Gd dendrite structures. Generally, specific patterns such as straight lines, corners, and "X" connections characterize edges. Traditional edge detection operators, such as the Sobel operator \cite{1973Pattern} and Robert operator \cite{do1961machine}, primarily identify edge information by computing gradient information such as pixel differences within the neighborhood. In contrast to general convolutional neural networks, where kernels are randomly initialized and not explicitly designed for edge detection, pixel-wise difference convolution is employed to enhance the extraction of Mg-Gd grain boundary information. The pixel-wise difference convolution-based deep learning model differs from general convolutional neural networks in that differences between pixel pairs are convolved instead of individual pixel values. This approach allows the model to better learn the variations between pixels. The Pidinet model is the first boundary detection network to utilize pixel-wise difference convolution for extracting edge features\cite{Su2021PixelDN}. By leveraging the principles of the Pidinet model and deep learning-based edge detection, a new edge detection model is constructed, as follows:
\begin{equation}
  \label{eq:model}
f:{X^{C*H*W}} \to {Y^{1*H*W}},
\end{equation}
where $C$, $H$ and $W$ represent the number of channels, height and width of the input image, respectively. The model $f$ denotes the edge detection model, whose learning objective is to minimize the error between the predicted results $X$ and the true labels $Y$. The specific network model is shown in Figure \ref{fig:process}.

\begin{figure*}[!htp]
  \centering
  \includegraphics[width=6in]{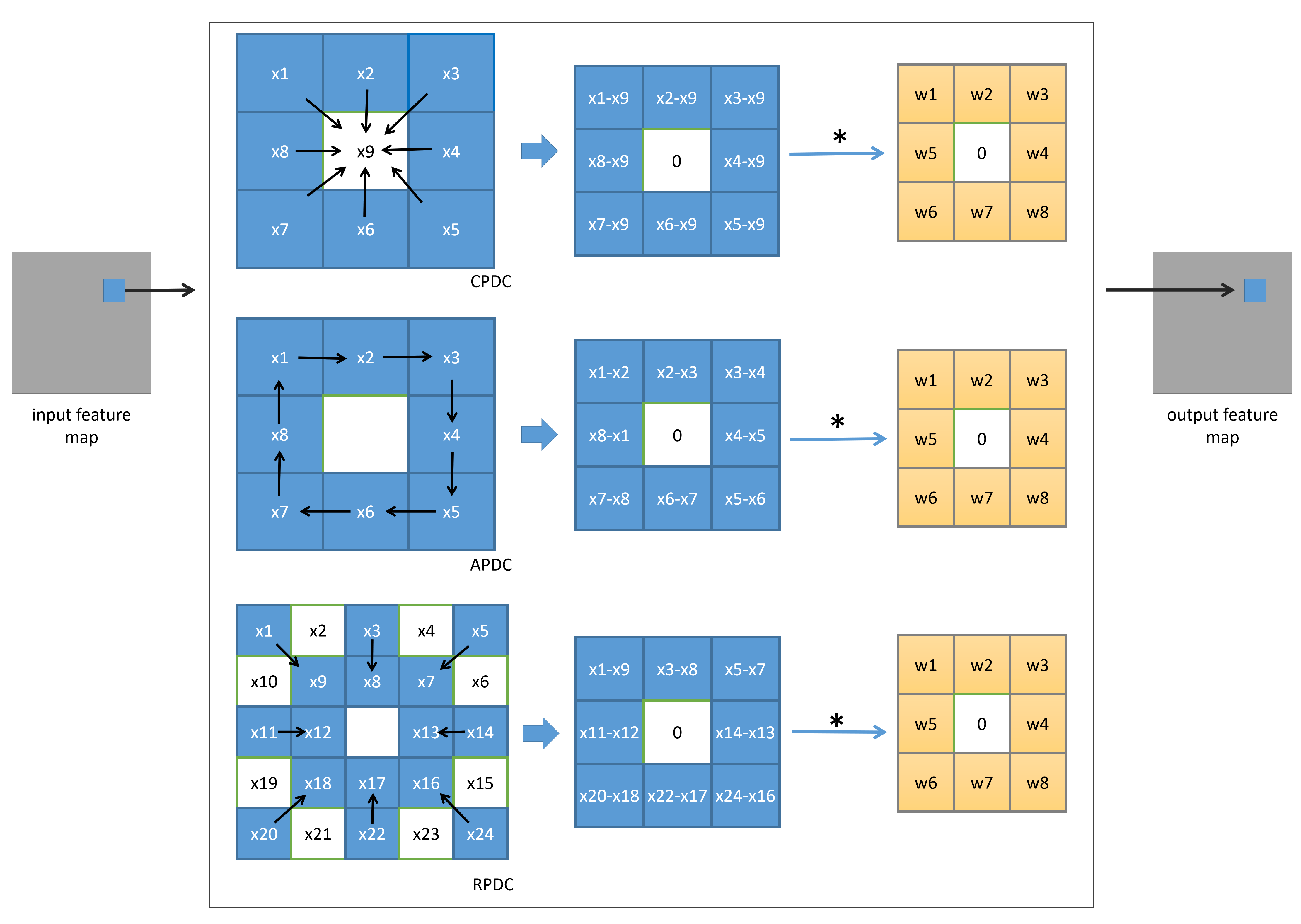}
  \caption{Three instances of pixel difference convolution derived from extended LBP descriptors. One can derive other instances by designing the picking strategy of the pixel pairs.}
  \label{fig:pixel}
\end{figure*}

When the model utilizes pixel-wise difference convolution \cite{Su2021PixelDN}, the original pixels in the local feature patch covered by the convolution kernel are substituted with pixel-wise differences. Three pixel-wise difference methods are applied, namely: Center Pixel-wise Difference in Local Feature (CPDC), Clockwise Pair-wise Difference (APDC), and the Difference between Outer and Inner Rings on a Larger Receptive Field (RPDC), as shown in Figure \ref{fig:pixel}. The architecture of the model consists of three stages, where each stage captures image features at different scales by incorporating the three pixel-wise difference methods. Each stage comprises four residual blocks, and the residual path in each block involves a sequence of pixel-wise difference convolution, ReLU layer, and 1*1 convolution. This configuration is applied iteratively.

Additionally, to further optimize the feature maps, a multi-scale adaptive feature refinement module(MSAFR) is connected after the last residual block of each stage, as shown in Figure \ref{fig:process}(b). This module is primarily composed of three main layers. The first layer is the Compact Part Convolution Module (CPCM)\cite{Su2021PixelDN}, designed to enrich multi-scale edge information, as shown in Figure \ref{fig:process}(c). This module takes the features extracted from the PDC structure of each stage as input, using four convolution kernels of different sizes and dilation rates to capture multi-scale features. By combining convolution kernels of various sizes with diverse dilation, this module can flexibly capture both local details and global information in the image. Larger kernels capture broader contextual information, while smaller kernels are better suited for capturing fine local details; the variation in dilation further extends the diversity of receptive fields. Thus, the CPCM module can effectively capture edge features at different scales, including fine edges and blurred contours, while retaining more contextual information from the image. Additionally, this module has significantly fewer output channels than input channels, effectively reducing the complexity of the model and addressing the limitation of traditional convolutions, which tend to focus solely on local feature information. The second introduced module is the Large Kernel Attention (LKA) module\cite{guo2023visual}. This module consists of two parts: the first layer is depth-wise convolution, used to extract local features by operating independently on each channel, thereby preserving the local edge details of each channel; the second layer is depth-wise dilation convolution, which combines broader contextual information to enhance the ability of the model to capture long-distance edges or large object contours, compensating for the limitations of depth-wise convolution in capturing global information, as shown in Figure \ref{fig:process}(d). The final layer is $1 \times 1$ convolution, used to fuse features across channels In the decoding process, the CPCM module uses multi-scale convolution kernels and dilation convolutions to help restore edge features at different scales, ensuring that both detailed edges and global contours are preserved. The LKA module further refines these multi-scale, multi-level features: depth-wise convolution handles local details, dilation convolution extends the receptive field to recover long-range dependencies, and $1 \times 1$ convolution fuses features across channels. Finally, the element-wise multiplication of features allows the model to adaptively enhance edge information, ensuring accuracy and robustness in edge detection.

\begin{figure*}[!htp]
  \centering
  \includegraphics[width=5in]{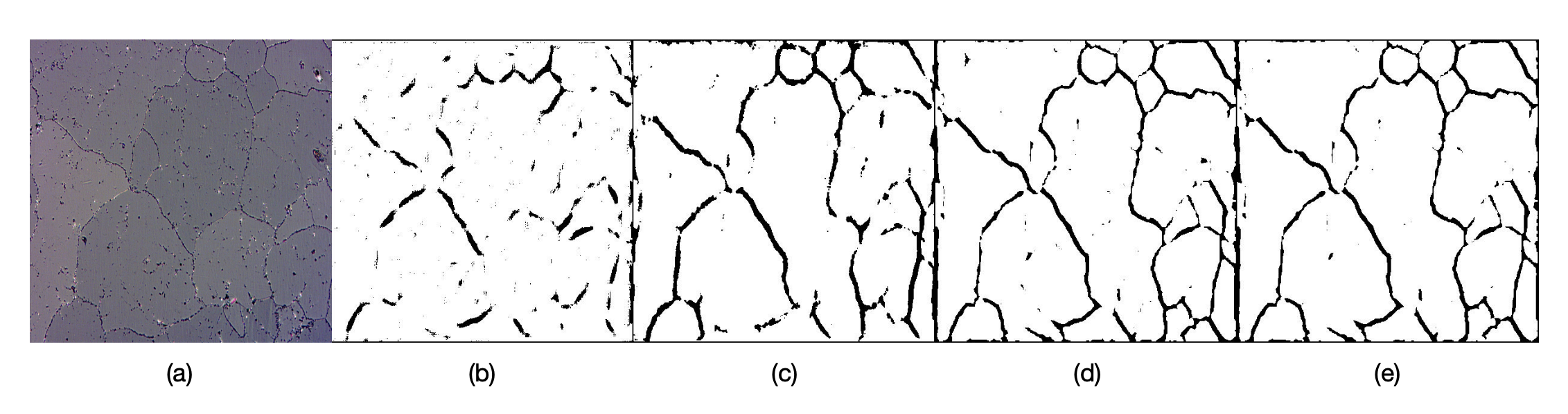}
  \caption{The edge detection model results at each stage are as follows: (a) represents the original image; (b) displays the grain boundary results detected in the first stage; (c) shows the grain boundary results detected in the second stage; (d) illustrates the grain boundary results detected in the third stage, and (e) presents the grain boundary results fused from the features of all three stages.}
  \label{fig:stagefeature}
\end{figure*}

Each stage further reduces the feature volume to a single-channel map through a $1 \times 1$ convolutional layer, which is then interpolated to the original size. The Sigmoid function is applied to create the edge map. The final edge map is obtained by fusing three single-channel feature maps through concatenation, convolutional layers, and the Sigmoid function. Like most deep learning-based edge detection models, the model primarily predicts the final result by fusing multi-scale and multi-level features extracted from the image. However, the setting of hierarchical levels in the model is not necessarily the more, the better, especially for images where the 
difference between edge features and background features is not particularly significant. Redundant model levels not only lead to the loss of useful features but also increase the runtime of the model. Additionally, to apply deep supervision at each stage, steering the detection results of each stage towards the direction of edge detection, the model is designed to incorporate deep supervision at each side output layer in every stage. Figure \ref{fig:stagefeature} illustrates the edge detection results at the output of each stage and the final fused result of this edge detection model.

Due to the significant class imbalance between the foreground and background in the Mg-2.1Gd alloy images, inspired by \cite{Li2019DiceLF}, for each generated edge map (including the final edge map), we utilized a loss function based on the Dice coefficient to optimize the model. The Dice coefficient is a metric ranging between 0 and 1, and our objective is to maximize it. The specific formulation is provided below:

\begin{equation}
  \label{eq:dice_loss}
Loss = \frac{{2\sum\limits_i^N {{p_i}{q_i}} }}{{\sum\limits_i^N {p_{_i}^2}  + \sum\limits_i^N {q_{_i}^2} }},
\end{equation}
where $i$ represents the pixel of the image, and $N$ is the number of pixels. $p_i$is the predicted value by the model, and $q_i$is the true value of the image.

\noindent\textbf{Edge Repair}

\begin{figure*}[!htp]
  \centering
  \includegraphics[width=6in]{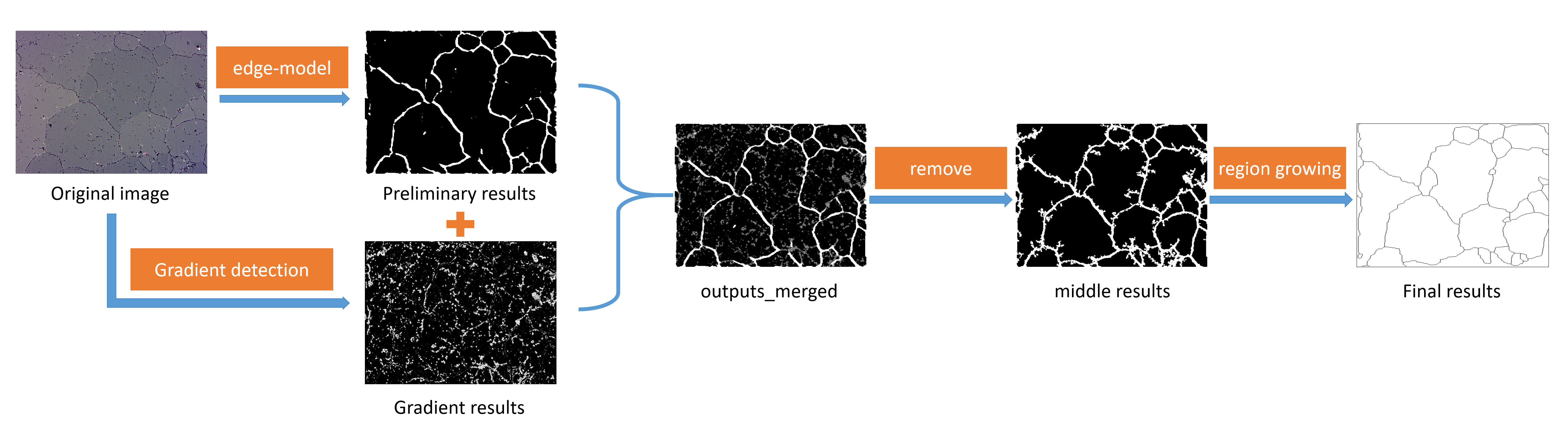}
  \caption{Detailed process of the edge repair module. The module utilizes image gradient information and edge detection structure to construct a mask, obtaining the final result using morphological methods from computer vision.}
  \label{fig:repair123}
\end{figure*}

The edge detection model is effective in extracting most grain boundaries. However, in the metallographic analysis of Mg-2.1Gd alloy images, the grayscale of weak boundaries is essentially the same as that of the internal regions of grains. Moreover, the overall sample quantity is sparse, making it challenging for the edge detection model to learn the feature differences of weak boundaries and internal grain regions. Besides weak boundaries, in the context of edge detection, the microstructure of the second phase precipitated in the alloy image serves as challenging noise. While the edge detection model can remove most of the noise, it tends to lose the grain boundaries connected to the noise during noise reduction. This leads to discontinuities in the detected grain boundaries, affecting the measurement of the average grain size, ultimately influencing the calculation of alloy yield strength and hardness. The edge restoration process can be roughly divided into two steps. The first step involves using the edge detection results and the gradient information of the image to restore fractured dendrite information. The second step employs morphology-based methods to eliminate false edge information, obtaining the final edge detection results. The specific process is illustrated in Figure \ref{fig:repair123}.

The gradient represents the direction and magnitude of the maximum rate of change at a specific point. In the context of images, regions with abrupt changes in pixel values correspond to edges and noisy areas. In our model, we employ image gradient information as a mask for edge restoration. The gradient information is combined with the edge detection results through addition, supplementing the undetected missing edges with gradient information to complete the preliminary stitching of the overall edges. Subsequently, the boundary auto-tracking algorithm \cite{Suzuki1985TopologicalSA} is employed to obtain the boundary information of the image, including position and area. By analyzing the topological structure of the binary image, noise information is removed, resulting in intermediate results. Finally, a region-growing method is applied to classify noise points and edge points in the repaired result image. This approach is primarily based on the similarity between pixels, merging pixels with similar properties to form connected regions. During the growth process, pixels are progressively merged into the same region according to predefined similarity criteria until further growth is not possible, The final result shown in Figure \ref{fig:repair123} is ultimately obtained.

\subsubsection{Extraction Method for Second Phase}

\begin{figure*}[!htp]
  \centering
  \includegraphics[width=6in]{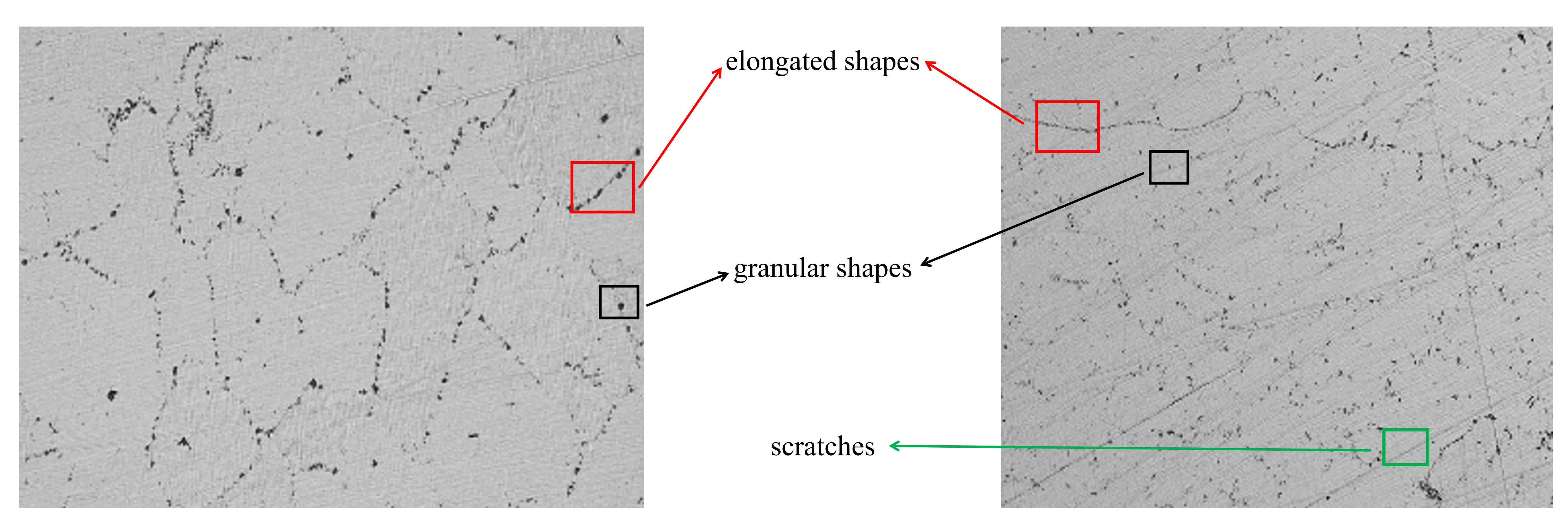}
  \caption{The microstructural image of the second phase, with the red box indicating the elongated second phase, the black box indicating the granular second phase, and the green box indicating scratches.}
  \label{fig:secondphase}
\end{figure*}

The morphology of the second phase in the Mg-2.1Gd alloy is characterized by both elongated and granular shapes, where the elongated form is composed of discontinuous granular second-phase particles. Although the second phase exhibits certain differences from the background features in the alloy image, during the alloy melting process, it is necessary to use sandpaper to polish the surface of the raw material to remove the oxide layer, resulting in scratches left on the alloy surface appearing in the alloy image, as shown in Figure \ref{fig:secondphase}. Traditional image processing methods, such as binarization, extract foreground information based on grayscale differences in images. However, in material images, the grayscale differences between the second phase and scratches are often very similar. As a result, binarization methods struggle to effectively distinguish between complex microstructural features and surface scratches. In contrast, deep learning methods offer significant advantages when processing images with intricate microstructures. We employ UNet++ as the model for extracting the microstructure of the second phase in Mg-Gd alloys. This model utilizes a multi-scale fusion strategy, which allows it to simultaneously learn both low-level detail features and high-level semantic information from the images. By combining and integrating features across different layers, it can more accurately capture the granular microstructure of the second phase and differentiate the continuous scratch texture features.

\begin{figure*}[!htp]
  \centering
  \includegraphics[width=6in]{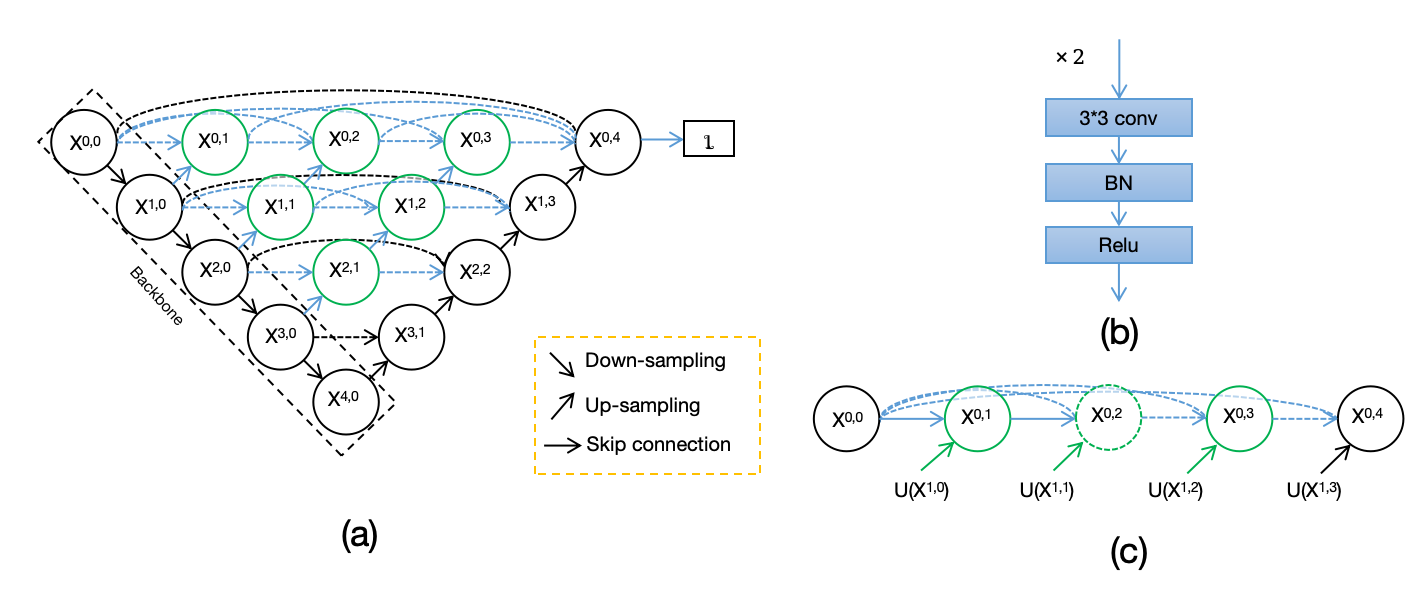}
  \caption{UNet++ model structure, where (a) represents the detailed framework of the UNet++ model; (b) illustrates the detailed structure of each convolution module; (c) showcases the UNet++ skip connection structure.}
  \label{fig:seconddetect}
\end{figure*}

Unet++ is an improved model based on the original Unet architecture, presenting an overall U-shaped structure \cite{8932614}. The left half represents the encoder path, undergoing four times downsampling, resulting in a total downsampling factor of 16. The right half constitutes the decoder path, symmetrically opposite to the left, and undergoes four times upsampling, as illustrated in Figure \ref{fig:seconddetect} (a). In the downsampling path, following the fundamental structure of convolutional neural networks, each downsampling step consists of two $3 \times 3$ convolutional units, with a BatchNorm layer and a ReLU activation layer following each convolutional unit, as depicted in Figure \ref{fig:seconddetect} (b). With each downsampling step, the model doubles the number of feature channels. In the upsampling path, each step involves upsampling a feature map and a ${\rm{2}} \times {\rm{2}}$ convolutional unit, reducing the number of feature channels by half. The resulting feature map is then concatenated with the corresponding feature map from the downsampling path. Unlike the original UNet model, UNet++ enhances segmentation accuracy by introducing Dense blocks and convolutional layers between the encoder and decoder of the model. Additionally, it incorporates skip pathways with new channels (depicted in green in Figure \ref{fig:seconddetect} (c)) to bridge semantic gaps between the encoder and decoder sub-paths. Furthermore, feature map concatenation occurs between every two consecutive convolutional blocks. The specific concatenation process is formulated as follows: 

\begin{equation}
  \label{eq:unet}
{x^{i,j}} = \left\{ \begin{array}{l}
  \begin{array}{*{20}{c}}
  {H({x^{i - 1,y}})}&{\begin{array}{*{20}{c}}
  {\begin{array}{*{20}{c}}
  {\begin{array}{*{20}{c}}
  {\begin{array}{*{20}{c}}
  {}&{}
  \end{array}}&{}
  \end{array}}&{}
  \end{array}}&{}
  \end{array}j = 0}
  \end{array}\\
  \begin{array}{*{20}{c}}
  {H([{x^{i,k}}]_{k = 0}^{j - 1},\mu ({x^{i + 1,j - 1}}))}&{j > 0}
  \end{array}
  \end{array} \right.,
\end{equation}
where $i$ and $j$ represent the numbers above the convolutional blocks in the figure, $\mu (x)$ denotes the bilinear interpolation upsampling operation, $[x, y]$ indicates channel concatenation of feature maps generated by two convolutional blocks x and y, and $H(x)$ represents two consecutive convolution operations. The specific structure of the skip pathways in the model is illustrated in Figure \ref{fig:seconddetect} (c). Each point in the horizontal direction is connected in the model to capture features at different levels. Moreover, different depth receptive fields exhibit varying sensitivity to targets of different sizes, with shallow layers being more sensitive to small targets and deep layers being more sensitive to large targets. After concatenation through the "concat" operation, these features are integrated to obtain image features from different receptive fields.

For semantic segmentation tasks, Binary Cross Entropy is widely used to measure the difference between the predictions of the model and the labels on the training data. In the case of binary classification tasks, it is expressed as:

\begin{equation}
  \label{eq:average-particle}
Loss_{BCE} = {y_n} \cdot \log {x_n} + (1 - {y_n}) \cdot \log (1 - {x_n}),
\end{equation}
where $y_n$ represents the label, and $x_n$ represents the predicted output.

\subsection{Microstructure Feature Parameters}

In the work, the three  microstructure feature parameters consist of secondary dendrite arm spacing (SDAS), the area fraction of the second-phase, and equivalent circle diameter (ECD) of the second-phase particles. Among them, SDAS and grain size are measured by the linear intercept method. Taking the SDAS for instance, it can be given as,

\begin{equation}
    \label{eq:sdas}
    SDAS = \frac{L}{n},
  \end{equation}
where $L$ is the length of the line drawn from edge to edge of the measured cells and $n$ is the number of dendrite cells. About 50 secondary dendrite arms are measured. For the microstructure analysis of the second phase in Mg-Gd alloy, an automated boundary tracking algorithm \cite{Suzuki1985TopologicalSA} is employed to extract boundary information, including location and area, by analyzing the topology of the binary image. The measurement method of the second-phase particles is as follows:

\begin{equation}
  \label{eq:radius}
 ECD = 2\sqrt {\frac{A}{\pi}},
\end{equation}
where $A$ represents the target area.

\subsection{Performance Evaluation}
Traditional experimental methods and empirical formulas often fall short in accurately revealing the complex relationships between composition, microstructural features, and properties in Mg-Gd alloys. To address this, we employ deep learning techniques to construct regression prediction models. In this study, we use the atomic percentage of Gd, grain size, second phase area fraction, and second phase particle size as input features, with Vickers hardness as the output feature. Prediction model as shown in Figure \ref{fig:transformerpPredict}. The prediction model is composed of an encoder and a decoder. First, the linear mapping layer of the encoder converts the input 4-dimensional features (including Gd atomic percentage, grain size, second-phase area fraction, and second-phase particle size) into a 64-dimensional high-level feature representation. Then, the self-attention mechanism is used to weight these features, where the $q$, $k$, and $v$ feature vectors are utilized for self-attention calculation. Specifically, the weights are obtained by taking the dot product of $q$ and $k$ \cite{vaswani2017attention}. The specific formulation is provided below:
\begin{equation}
  \label{eq:attention}
  Attn(q,k,v) = soft\max (q{k^T}/\sqrt M )v,
\end{equation}
where softmax is the activation function, and $1/\sqrt{M} $ is the scaling factor, which ultimately weights the feature vector $v$. Through the multi-head attention mechanism (with 4 attention heads), the model is able to capture diverse features from different perspectives within the data. To extract sufficiently complex features from a small dataset while preventing model overfitting, we chose a 3-layer Transformer module. This ensures the expressive capability of the model while also reducing its complexity. In the decoder part of the prediction model, a fully connected (FC) layer is used for feature decoding. In the FC layer, the number of neurons in the input layer corresponds to the dimensionality of the input feature vector, and the output value $y$, represents the estimated Vickers hardness.

We use leave-one-out cross-validation(LOO-CV) to train the models on the dataset of feature data from the Mg-Gd alloys. This method ensures that each sample appears in the validation set exactly once, providing a robust assessment of the generalization capabilities of the models. In this study, we use the Mean Squared Error (MSE) loss function to optimize the predictive accuracy of our deep learning models. The MSE loss function is widely used in regression tasks because it measures the difference between predicted and actual values and is more sensitive to larger errors. By minimizing the MSE loss function, we aim to make the predicted values as close as possible to the actual values, thereby improving the accuracy of the model and ensuring that the models effectively capture the relationship between the input features and Vickers hardness. Meanabsolute error (MAE), mean squared error (MSE), root meansquared error (RMSE), and coefficient of determination (R2)are used to evaluate the model accuracy. They are shown as follows:
\begin{equation}
  \label{eq:MAE}
  MAE = \frac{1}{n}\sum\limits_{i = 1}^n {\left| {{y_i} - {{\widehat y}_i}} \right|},
\end{equation}
\begin{equation}
  \label{eq:MSE}
  MSE = \frac{1}{n}\sum\limits_{i = 1}^n {{{({y_i} - {{\widehat y}_i})}^2}},
\end{equation}
\begin{equation}
  \label{eq:RMSE}
  RMSE = \sqrt {\frac{1}{n}\sum\limits_{i = 1}^n {{{({y_i} - {{\widehat y}_i})}^2}} },
\end{equation}
\begin{equation}
  \label{eq:R2}
  {R^2} = \frac{{\sum\limits_{i = 1}^n {{{({y_i} - {{\widehat y}_i})}^2}} }}{{\sum\limits_{i = 1}^n {{{({y_i} - {{\overline y }_i})}^2}} }},
\end{equation}
\begin{figure*}[!htp]
  \centering
  \includegraphics[width=2in]{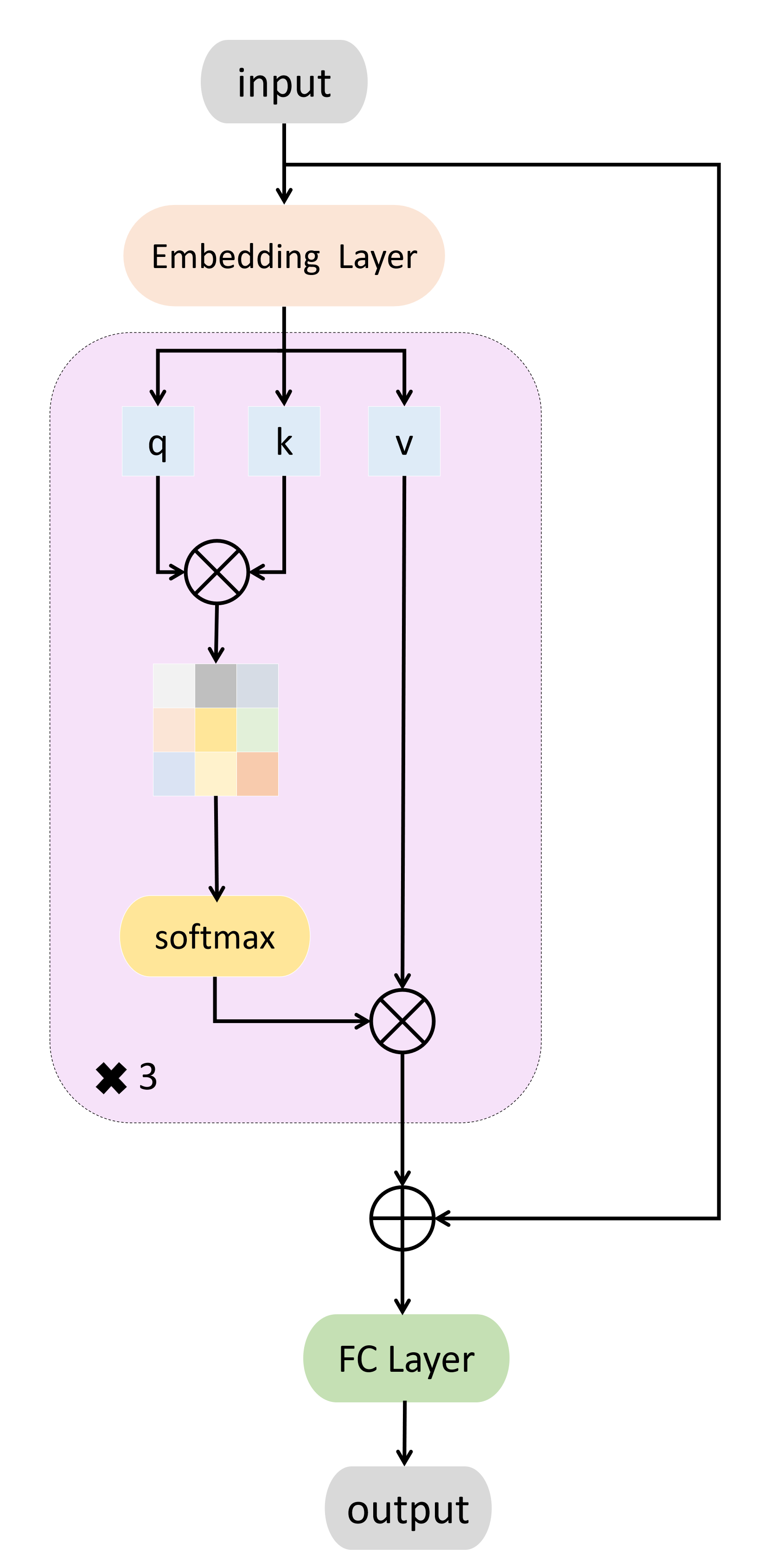}
  \caption{Structure of the regression prediction model based on Transformer.}
  \label{fig:edgeResult}
\end{figure*}
where $y_i$ is the actual value of the i-th sample, ${\widehat y}_i$ is the predicted value of the i-th sample, ${\overline y }_i$ is the mean of all label ${\widehat y}_i$. The MAE isthe average of the absolute differences between predictedand true values. Smaller MAE values indicate better accuracy. The MSE is the average of the squared differencesbetween predicted and true values. The smaller the MSE value, the better the accuracy of the model predictions. The RMSE is the arithmetic square root of the MSE,which is used to measure the deviation between the predictedvalue and the true value. The $R^2$ is a measure of the per-formance of the regression model. The closer the $R^2$ is to 1, the better will be the performance of the regression model.

\section{Experiments and Results}
\label{sec:experiments}
\subsection{Data}

Our experimental data includes nine types of Mg-2.1Gd alloy samples from experiments, as well as twelve types of Mg-Gd alloy data from the literature, specifically Mg-(0.08, 0.16, 0.31, 0.49, 0.52, 1.67, 2.06, 2.43, 2.45, 2.46, 2.49, 2.9)Gd. A total of 67 images were utilized, with 44 images for training the microstructural extraction model and 13 images for testing the model. Due to the significant imbalance between grain boundary information and background information in the alloy images, we employed data augmentation techniques such as rotation, flipping, and cropping. This expanded the grain boundary detection training set from the original 44 images to 11,862 images. This approach helps improve the accuracy and robustness of the model in detecting grain boundaries across different images, thereby enhancing the quality of microstructural analysis of the alloy materials. In the material property prediction phase, 19 groups of Mg-Gd alloy data were used for training and 2 groups for testing. These experimental data encompass all the necessary information required for training and testing both the microstructural extraction model and the material property prediction model, ensuring the effectiveness and stability of the models.

\subsection{Experiments Results}
For the microstructure metallographic images of Mg-Gd alloy, we used evaluation standards based on edge detection and semantic segmentation to assess the detection results for both types of images. The evaluation metrics for edge detection results include precision, recall, and F-score to assess the performance of the detection algorithm. Precision, an evaluation metric, is used to assess the probability that the pixels identified by the model as grain boundaries are indeed true grain boundary pixels. The higher the precision, the better the edge detection performance. Recall, an indicator, represents the proportion of detected true boundary pixels out of all true boundary pixels. The higher the recall, the more true edges the edge detection model captures. F-score, the harmonic mean of precision and recall.

\begin{figure*}[!htp]
  \centering
  \includegraphics[width=6in]{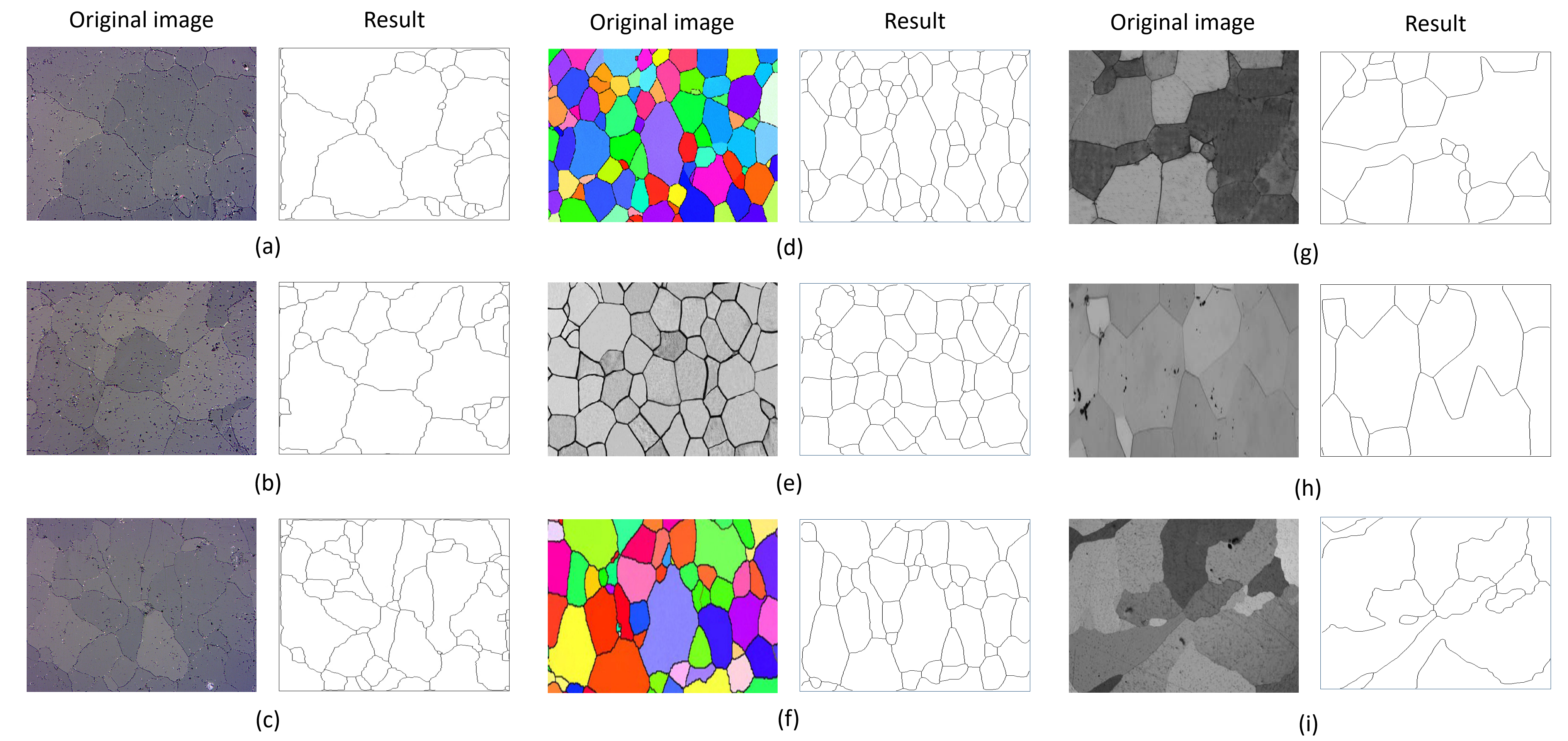}
  \caption{Grain boundary detection results for Mg-Gd solid solution alloys are as follows: (a-c) Images of Mg-2.1Gd (at.$\%$) alloy; (d) Image of Mg-2.9 (at.$\%$) Gd alloy; (e) Image of Mg-0.49 (at.$\%$) Gd alloy; (f) Image of Mg-1.67 (at.$\%$) Gd alloy; (g) Image of Mg-2.46 (at.$\%$) Gd alloy; (h) Image of Mg-2.45 (at.$\%$) Gd alloy; (i) Image of Mg-0.16 (at.$\%$) Gd alloy.}
  \label{fig:edgeResult}
\end{figure*}

\begin{table*}[h!]
  \begin{center}
    \caption{The comparison of mean of each method evaluated on Mg-Gd alloy image edge detection results.}
    \label{tab:evaluation_edge}
    \resizebox{7cm}{1cm}{
    \begin{tabular}{cccc}   
      \hline
                    & Precision            & Recall            & F1-score                  \\ \hline
      RCF           & 0.5663	             & 0.7416	           & 0.6422                 \\\hline
      BDCN          & 0.3571	             & 0.5903	           & 0.4450          \\ \hline
      Our Method    & 0.8391	             & 0.6613	           & 0.7391          \\ \hline
    \end{tabular}}
  \end{center}
\end{table*}

For edge detection data, there is very little effective information that can be gleaned from each image. If the difference in grayscale between the image edge and non-edge areas is not significant, the difficulty of edge detection will be greatly increased. The edge detection model used in this study effectively extracts edge information by learning the differences between each pixel and its surrounding pixels through convolution of class pixel differences. By incorporating only three stages to maintain weak edge features while removing noise, the model successfully repairs broken edges, yielding optimal results. Simultaneously, to verify the effectiveness of the model used, we compare the deep learning-based edge detection models RCF, BDCN, and our proposed model using rating metrics. The edge detection results obtained by our proposed model are optimal, and the specific evaluation indicators are shown in Table \ref{tab:evaluation_edge}.

\begin{figure*}[!htp]
  \centering
  \includegraphics[width=6in]{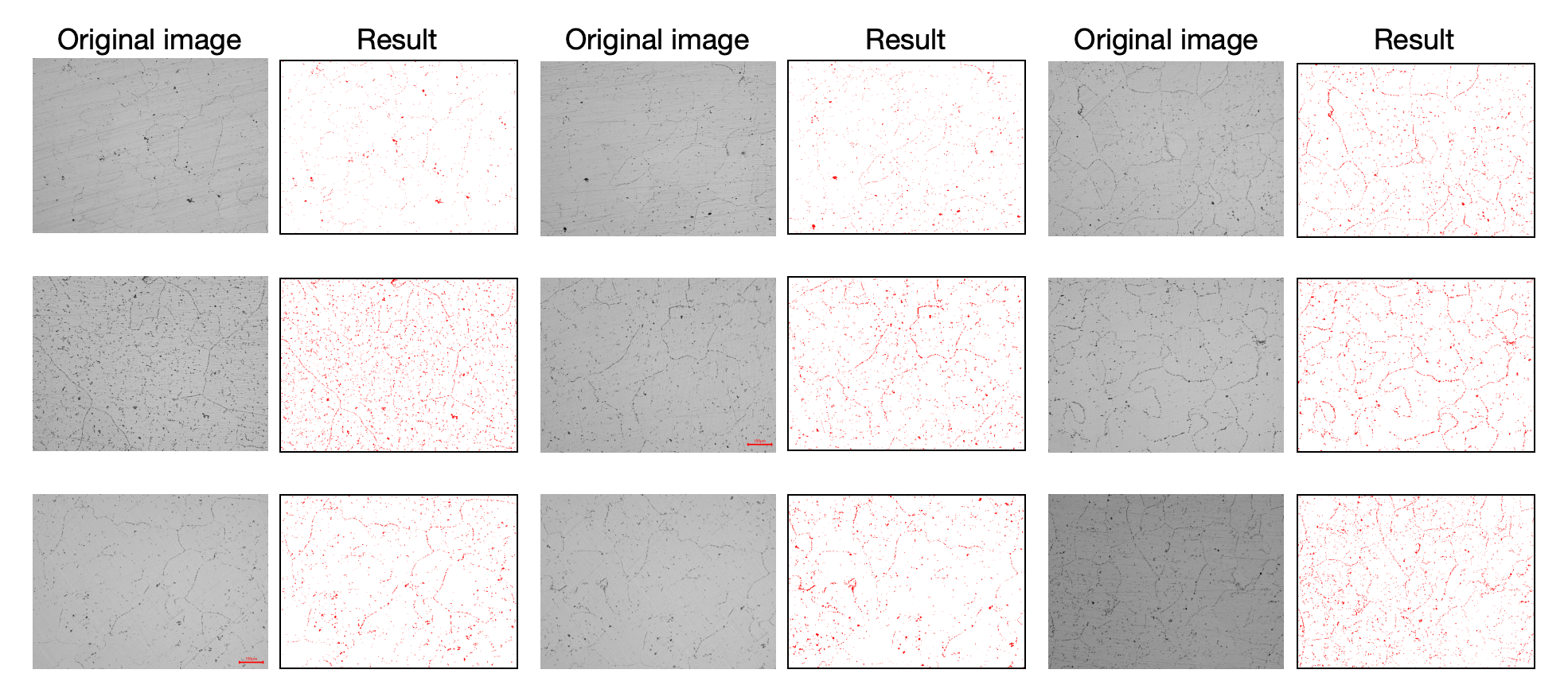}
  \caption{Original image of the second phase in Mg-Gd alloy and semantic segmentation results.}
  \label{fig:secondresult}
\end{figure*}

\begin{table*}[h!]
  \begin{center}
    \caption{The comparison of mean of each method evaluated on Mg-Gd alloy images second phase segmentation results.}
    \label{tab:evaluationimg2}
    \resizebox{7cm}{1cm}{
    \begin{tabular}{cccc}   
      \hline
                    & Accuracy             & Precision         & MIOU                  \\ \hline
      FCN           & 0.9656               & 0.6382            & 0.9656                 \\\hline
      UCTransNet    & 0.9484	             & 0.32375	         & 0.9725          \\ \hline
      UNet++        & 0.9949	             & 0.8313	           & 0.9786          \\ \hline
    \end{tabular}}
  \end{center}
\end{table*}

By comparing classical and state-of-the-art semantic segmentation models, it is evident that UNet++ produces optimal results, as shown in Table \ref{tab:evaluationimg2}. For semantic segmentation models, as the network deepens, information loss occurs through consecutive down-sampling operations. UNet++ addresses the shortcomings of conventional models by progressively capturing features at different levels during the down-sampling process. In the feature fusion phase of up-sampling, it integrates features with larger receptive fields and those with smaller receptive fields, effectively combining detailed feature information with positional feature information. This enhances the accuracy of positional information without compromising precision, resulting in an overall improvement in segmentation accuracy. Refer to Figures \ref{fig:secondresult} for illustrative semantic segmentation results.

\subsection{The Relationship between Microstructure and Material Properties.}

\begin{figure*}[!htp]
  \centering
  \resizebox{14cm}{14cm}{\includegraphics{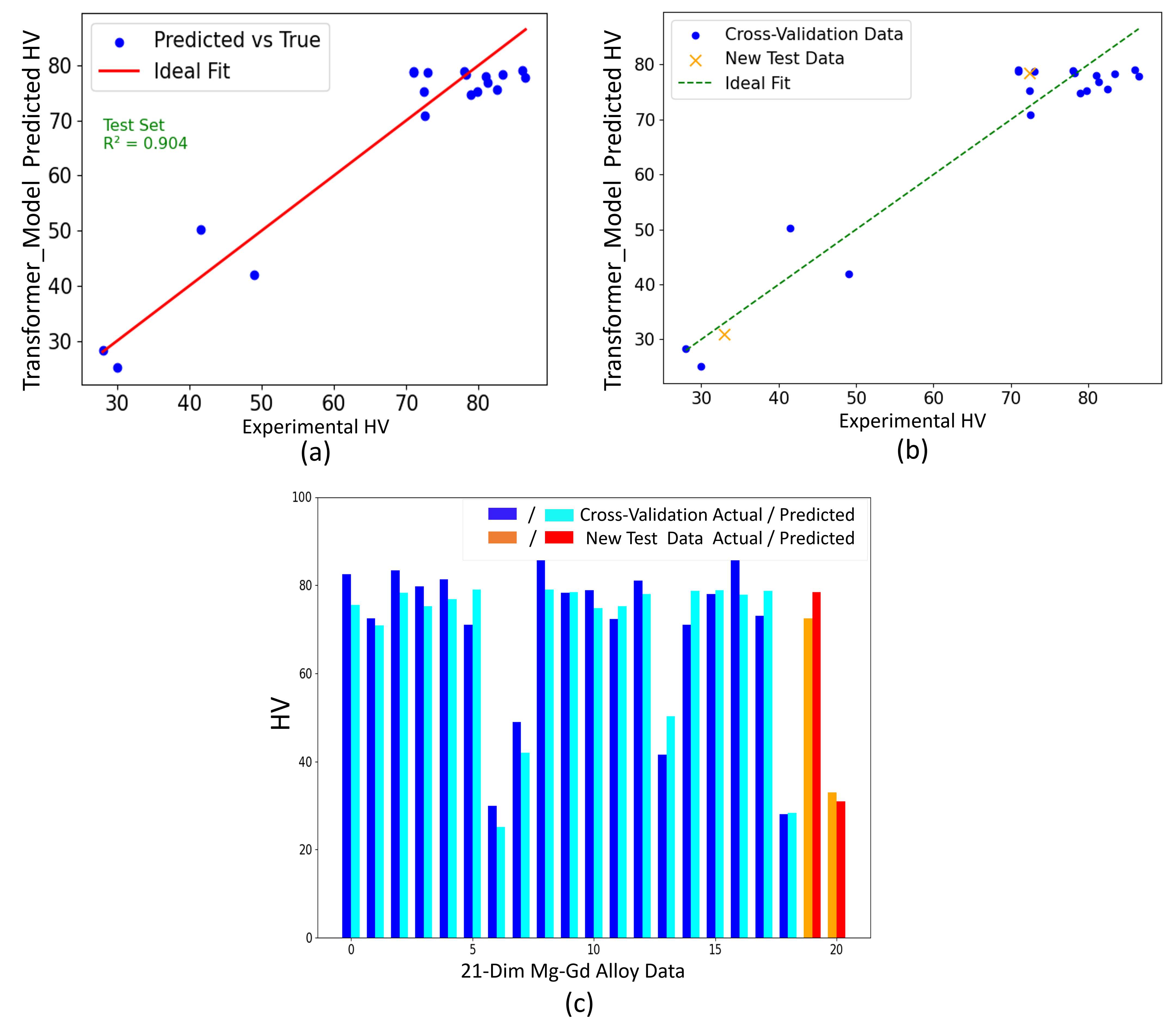}}
  \caption{The performance of the Transformer model in predicting Vickers hardness: (a)comparison of the predicted results of the best model (Transformer) with the experimental results on the test sets, (b)the regression scatter plot of the actual and predicted values for the 21 data groups, (c)bar chart comparing the errors between the actual and predicted values for 21 data sets, respectively. HV, vickers-hardness.}
  \label{fig:TransformerModel}
\end{figure*}

The experimental results, shown in Figure \ref{fig:TransformerModel}, demonstrate the excellent performance of the 3-layer regression model constructed using Transformer in predicting the Vickers hardness of Mg-Gd alloys. In the regression plot of Figure \ref{fig:TransformerModel}(b), most cross-validation (blue dots) and test set (orange crosses) data points are close to the ideal fitting line, indicating strong model performance, particularly in the higher hardness range (70-80 HV). Despite some deviations in the lower hardness range (30-50 HV), the error remains within acceptable limits. Meanwhile, in the bar chart of Figure \ref{fig:TransformerModel}(c), the difference between actual and predicted values for cross-validation data (blue and cyan bars) is minimal, highlighting strong fitting ability during training. While some discrepancies are observed in samples such as group 5 and group 16, the overall error is small. In the test set (red and orange bars), the predicted values align closely with the actual values, exhibiting minimal error, particularly for group 21, where the error is almost negligible. Specifically, group 20 has a Gd atomic percentage of 2.1 at$\%$, with a predicted Vickers hardness of 78.43, compared to an experimental hardness of 72.42, resulting in an error of 6.01. The predicted hardness for this group is slightly higher than the actual value, which may be attributed to feature diversity or noise in the dataset. In contrast, group 21 has a Gd atomic percentage of 0.31 at$\%$, with a predicted hardness of 30.95, while the experimental hardness is 33, leading to an error of only 2.05.

\begin{figure*}[!htp]
  \centering
  \resizebox{14cm}{14cm}{\includegraphics{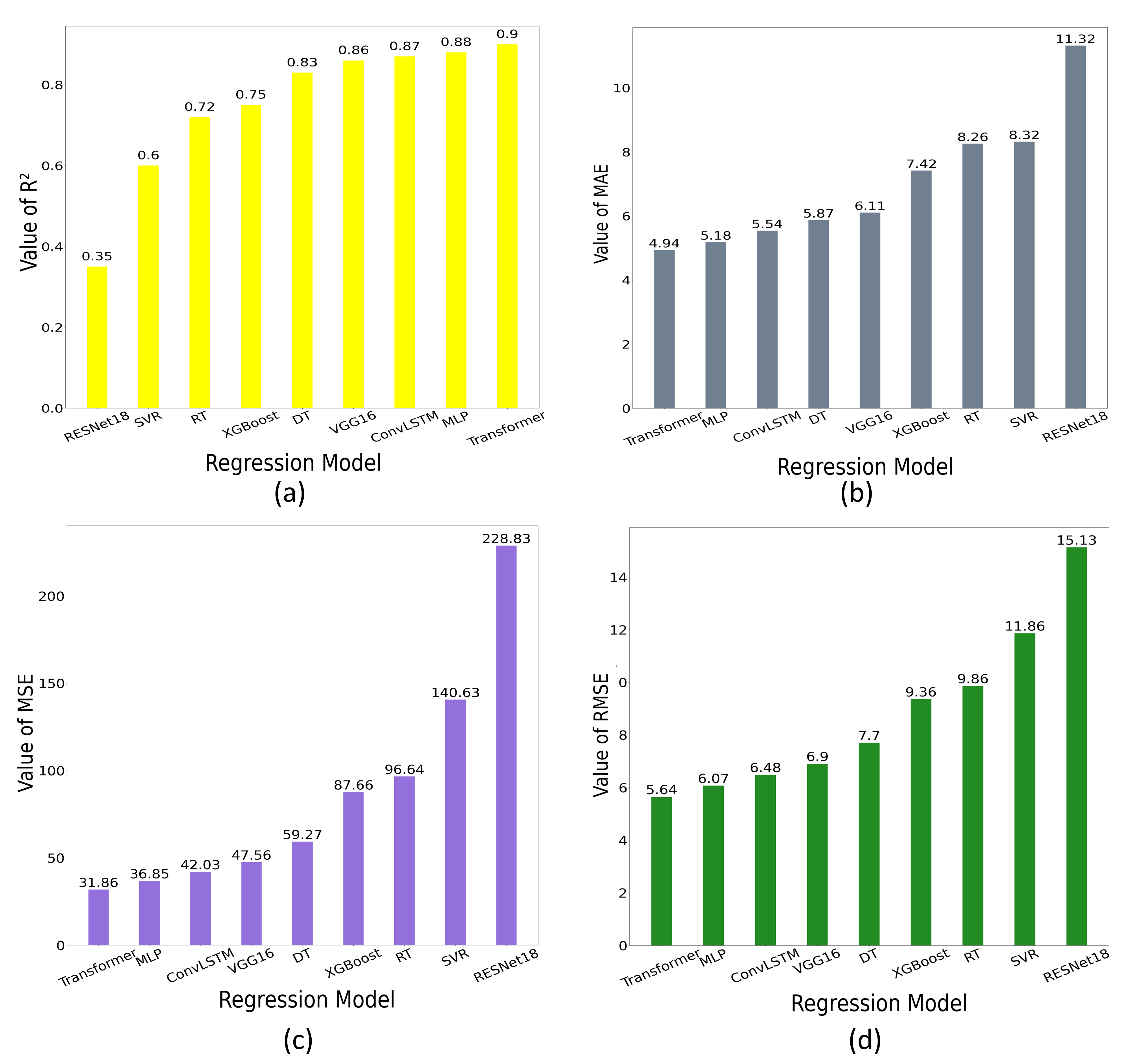}}
  \caption{Model performance of different machine learning and deep learning algorithms on predicting the HV:  (a) $R^2$, (b) MAE, (c) MSE, (d) RMSE, respectively. MAE, meanabsolute error; MSE, mean squared error; RMSE, root mean squared error.}
  \label{fig:regressionModel}
\end{figure*}

To further validate the effectiveness of the model, the regression model constructed based on Transformer was compared with eight other models, including four deep learning and four machine learning algorithms, such as ConvLSTM time series model \cite{NIPS2015_07563a3f}, Multilayer Perceptron (MLP) \cite{NEURIPS2019_bdbca288}, ResNet18\cite{He_2016_CVPR}, VGG19 \cite{simonyan2014very}, Decision Tree(DT) \cite{quinlan1986induction}, Random Forest(RF) \cite{breiman2001random}, XGBoost \cite{10.1145/2939672.2939785} and Support Vector Regression (SVR) \cite{MOHANDES2004939}. The performance of each model is shown in the Figure \ref{fig:regressionModel}. The results indicate that the Transformer regression deep learning model constructed in this study performs the best on the test set, achieving MAE, MSE, RMSE, and $R^2$ values of 4.94, 31.86, 5.64, and 0.9, respectively. To prevent overfitting, leave-one-out cross-validation was applied to the dataset. Overall, the multi-layer self-attention mechanism of the Transformer model effectively captures the complex relationship between Mg-Gd alloy microstructure and Vickers hardness, showing excellent fitting and generalization capabilities even with small sample datasets.

\begin{figure*}[!htp]
  \centering
  \includegraphics[width=6in]{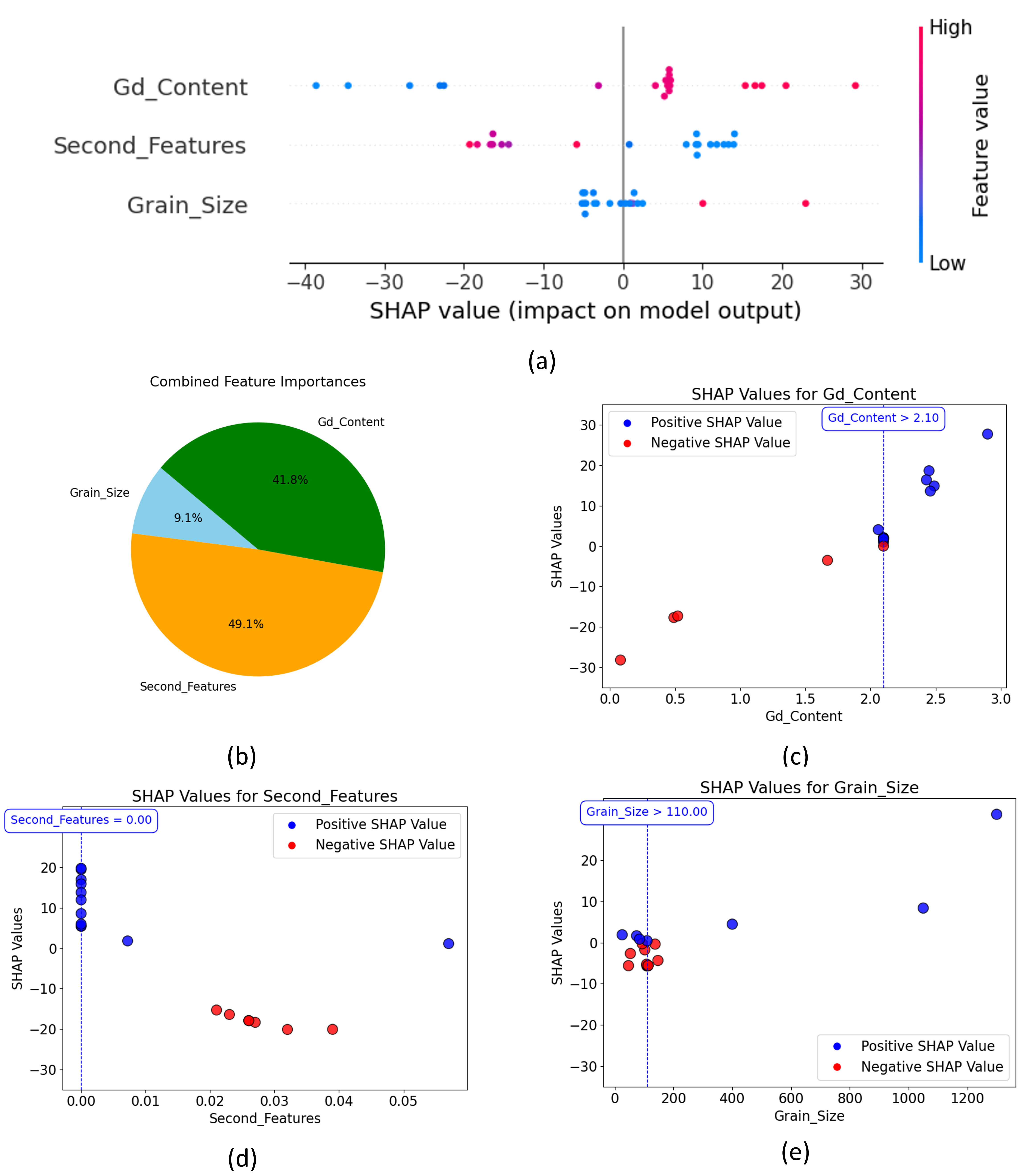}
  \caption{(a) Summary plot of SHAP values, revealing the individual feature contributions to the model predictions; (b) Feature importance of the MLP model; (c-e) SHAP values (positive in blue, negative in red) of Gd content (c), Second phase feature (d), Grain size(e); SHAP, Shapley additiveexplanations.}
  \label{fig:regressionShap}
\end{figure*}

Compared to traditional methods for predicting material properties, such as empirical formulas and multimodal learning models, the approach used in this study, which is based on composition and alloy microstructural features, offers broader applicability and better interpretability. The limitation of empirical formulas is that they can only be applied to specific materials or conditions, and some parameters in the formulas lack fixed values, requiring adjustment based on specific conditions, which can lead to inaccurate predictions. On the other hand, multimodal learning models often involve convolutional neural networks (CNNs) to extract microstructural features from images, which can be challenging to interpret. In contrast, microstructural feature parameters are more strongly correlated with performance and can better reveal the relationships between alloy composition, microstructure, and material properties. To assess the contribution of each feature to the HV regression prediction model, the feature importance of the optimal three layers transformer model was calculated, as shown in Figure \ref{fig:regressionShap}(b). The ranking of feature importance clearly identifies that the Gd atomic percentage has the highest influence, followed by second-phase characteristics, including second-phase area fraction and particle size, while the average grain size has the least impact on the Vickers hardness of Mg-Gd alloys.

To further explain how the actual values of features affect HV, SHAP values were calculated using the SHAP method to measure the contribution of each feature to the predicted value. Each feature value corresponds to a Shapley value, where positive or negative Shapley values indicate an increase or decrease in the prediction of the model, respectively. Figure \ref{fig:regressionShap}(a) shows a summary plot of SHAP values, where the x position of the dot represents the impact of the feature value on the model prediction, and the color of the dot indicates the magnitude of the feature value. The significant features affecting the HV regression prediction model are ranked as Gd atomic percentage, second-phase features, and average grain size, consistent with the feature importance analysis results. Feature importance analysis typically evaluates the overall model performance, while SHAP analysis provides local interpretative evaluations for each sample. The combination of these methods confirms the accuracy of the results regarding the features with substantial impact on the model.

Figure \ref{fig:regressionShap} presents the SHAP values for significant features and their respective feature values. The figures reveal several important critical values, where SHAP values on either side of these critical values are primarily positive (blue dots) or negative (red dots). In Figure \ref{fig:regressionShap}(c), it can be observed that when the Gd atomic percentage exceeds 2.1 at$\%$, the corresponding SHAP values are positive, indicating an increasing trend in HV. Similarly, in Figure \ref{fig:regressionShap}(d), when the second-phase fraction approaches 0, HV also increases. Given that the dataset used for the regression prediction model includes 8 types of data with a Gd content of 2.1$\%$ and 12 types with other Gd contents, with a higher proportion of 2.1$\%$ data, the model tends to identify 2.1$\%$ as a critical value during SHAP analysis. Specifically, the large proportion of data with 2.1$\%$ Gd content causes the model to focus on this composition, establishing it as a key parameter affecting Vickers hardness (HV). The SHAP analysis shows that when Gd content exceeds 2.1$\%$, the predicted HV increases, indicating that 2.1$\%$ is a valuable reference value for practical applications. In future alloy design, controlling the Gd content around 2.1$\%$ could optimize the hardness performance of Mg-Gd alloys. This approach not only improves the mechanical performance of the alloy but also provides reliable data support and theoretical foundations for new material development. This finding offers new perspectives for the composition control and performance optimization of Mg-Gd alloys and further validates the potential of machine learning models in material science.

In summary, using 2.1$\%$ as a reference value for Gd content can enhance the efficiency of alloy design and ensure performance stability in practical applications. This research outcome holds significant guidance for the development and application of Mg-Gd alloys in the future.

\section{Conclusion} 
\label{sec:conclusion}

This study establishes  a comprehensive research model for solid solution Mg-Gd alloys, utilizing deep learning methods to efficiently extract microstructural features from various Mg-Gd alloy images. By combining the Gd atomic percentage of Mg-Gd solid solution alloys with three key microstructural features, an accurate Vickers hardness (HV) prediction model for Mg-Gd alloys was constructed using a Transformer model. Compared to traditional manual extraction methods, this approach avoids human interference, significantly saves time and effort, and offers higher accuracy than general-purpose materials software like ImageJ. This automated and precise feature extraction method provides researchers with an efficient and reliable tool for analyzing and predicting the performance of alloy materials. The proposed framework successfully establishes the mapping relationships among material composition, microstructure, and performance, offering new insights and methods that could potentially be applied to research on other alloy materials research.

Through feature importance analysis and SHAP (Shapley Additive Explanations) analysis, we identified four key features (including composition and microstructure) that significantly impact the Vickers hardness of Mg-Gd alloys, along with their critical values. These features include the Gd atomic percentage, average grain size, second phase area fraction, and second phase particle size. The study results indicate that the Gd atomic percentage is the most critical feature affecting alloy hardness, while the average grain size has the least impact. These findings not only reveal the primary factors influencing alloy hardness but also provide specific guidance for optimizing alloy composition and microstructure. It is noteworthy that this study used a dataset of only 20 types of Mg-Gd alloy, which is considered a small dataset. Despite the limited data, the method achieved high prediction accuracy, demonstrating its superiority and reliability with small datasets.

In the future, we can expand the scale and diversity of the dataset by introducing more varieties of Mg-Gd alloy samples and data from different experimental conditions to enhance the generalization ability and prediction accuracy of the model. Additionally, applying this method to other types of alloys will validate its broad applicability. By integrating cross-disciplinary data, we can further enhance the generality and practicality of the model.

\bibliographystyle{unsrt}  


\end{document}